\documentclass{CVIS}
\usepackage{amsmath,bm}
\usepackage{graphicx}
\usepackage{dcolumn}
\usepackage{subcaption,soul}
\usepackage{caption}
\usepackage[numbers,sort&compress,sectionbib]{natbib}
\usepackage[pagebackref=flase,breaklinks=true,letterpaper=true,colorlinks=true,linkcolor = blue, urlcolor = black,citecolor=black,bookmarks=true]{hyperref}
\usepackage{url}
\usepackage{mleftright}
\usepackage{multirow}

\usepackage{titlesec}
\usepackage[redeflists]{IEEEtrantools}

\title{Performance or Trust? Why Not Both. Deep AUC Maximization with Self-Supervised Learning for COVID-19 Chest X-ray Classifications}

\begin{document}
\bstctlcite{IEEEexample:BSTcontrol}
\sloppy
\author{
\begin{tabularx}{\textwidth}{X X}
Siyuan He & National Research Council Canada\\
Pengcheng Xi  & National Research Council Canada\\
Ashkan Ebadi  & National Research Council Canada\\
St\'{e}phane Tremblay  & National Research Council Canada\\
Alexander Wong  & University of Waterloo\\
\multicolumn{2}{l}{Email: \{\textit{Firstname.Lastname}\}@nrc-cnrc.gc.ca; a28wong@uwaterloo.ca}\\
\end{tabularx}
}

\maketitle
\begin{abstract}
	Effective representation learning is the key in improving model performance for medical image analysis. In training deep learning models, a compromise often must be made between performance and trust, both of which are essential for medical applications. Moreover, models optimized with cross-entropy loss tend to suffer from unwarranted overconfidence in the majority class and over-cautiousness in the minority class. In this work, we integrate a new surrogate loss with self-supervised learning for computer-aided screening of COVID-19 patients using radiography images. In addition, we adopt a new quantification score to measure a model's trustworthiness. Ablation study is conducted for both the performance and the trust on feature learning methods and loss functions. Comparisons show that leveraging the new surrogate loss on self-supervised models can produce label-efficient networks that are both high-performing and trustworthy.

\end{abstract}

\section{Introduction}

COVID-19 continues to affect our daily lives. In the fight against the pandemic, computer-aided screening of patients using radiography images has served as a complementary approach to standard polymerase chain reaction (PCR) test. Despite our efforts in improving the performance of deep learning models \cite{zhang2021covid19, as2020covid}, a compromise often must be made between the performance and model trust \cite{wong2021really}. \\



Effective representation learning is the key in improving model performance. The most common approach is through supervised learning on large-scale data sets with labels. It can also be realized through un-supervised training, when data labels are missing or expensive to collect. A series of self-supervised models have achieved comparable performance to its supervised counterparts on benchmark data sets \cite{He_2020_CVPR, 49372}. They learn image representations through minimizing an embedding distance between image pairs derived from the same image, while maximizing the distance between the pairs from different images. \\

Regarding model trust, a deep classification neural network optimized on cross-entropy loss tends to be over-confident in its incorrect predictions of the majority class, and overcautious in its correct predictions of the minority class \cite{doi:10.1021/acs.jcim.1c00160}. In this work, we investigate a new surrogate loss function termed deep AUC Maximization \cite{yuan2021robust}. We integrate it with self-supervised models pre-trained with the MoCo framework \cite{He_2020_CVPR} for improving model performance and trust. \\

To our best knowledge, this is the first integration of the deep AUC Maximization loss function with self-supervised learning. In addition, we validate the models through quantitative comparisons to gain insight into the models’ trustworthiness. Our assumption is that, by adopting the new surrogate loss function to self-supervised models, we no longer need to sacrifice model trust for performance but can achieve both. In summary, our contributions are threefold:


\begin{itemize}
    \item We proposed the use of a new surrogate loss on self-supervised models to improve representation learning and model trust pertinent to the task of screening COVID-19 patients with deep learning;
    \item We demonstrated with experimental results that the self-supervised models had improved model performance over supervised ones on screening COVID-19 patients;
    \item We showed that the use of a new surrogate loss can produce models that are more trustworthy than those optimized with cross-entropy loss.
\end{itemize}


\section{Literature Review}

Self-supervised learning has gained momentum in learning visual representations. It can be categorized into generative and discriminative approaches. As a discriminative method, Momentum Contrast (MoCo) trains a visual representation encoder by matching an encoded query to a dictionary of encoded keys through a contrastive loss \cite{He_2020_CVPR}. The query encoder is shared with the key encoder, which receives slow updates in order to achieve consistency in learning visual representations.\\


In medical AI, contrastive learning has led to improved representation learning. In \cite{pmlr-v143-sowrirajan21a}, a model named MoCo-CXR proved that linear models trained on MoCo-CXR-pretrained representations outperform those without MoCo-CXR-pretrained representations. Due to the scarcity of COVID-19 patient data, the MoCo model has been applied to predicting patient deterioration based on chest X-rays \cite{sriram2021covid}.\\

Area under the Receiver Operating Characteristic curve (AUC) is widely used in medical image analysis for evaluating the performance of a neural network. Recently, Yuan \textit{et al.} \cite{yuan2021robust} proposed a novel surrogate loss over the standard cross-entropy loss to directly optimize for the AUC metric. AUC maximization, as the authors claim, can lead to the largest increase in a network's performance. This new surrogate loss function was integrated with supervised deep learning models and it achieved the first place in the Stanford CheXpert competition \cite{yuan2021robust}.

\section{Methodology}

\subsection{Model Architecture}

\begin{figure*}[h!]
	\begin{center}
		\includegraphics[scale = 0.4]{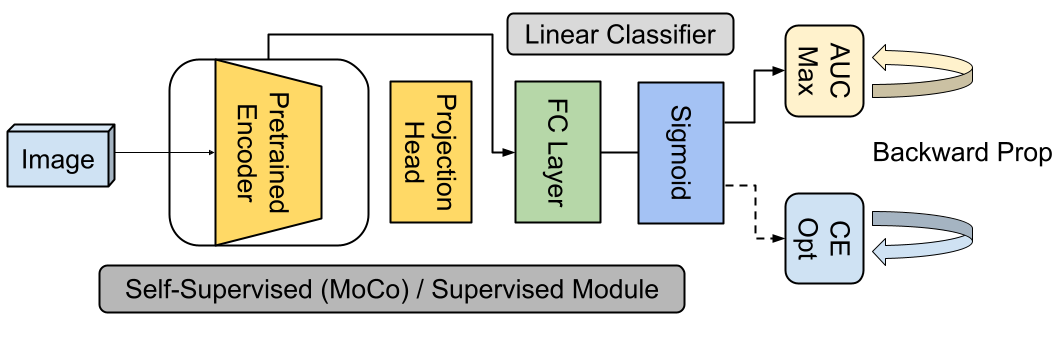}
	\end{center}
	\caption{Model Architecture and Fine-tuning Pipeline.}
	\label{fig2}
\end{figure*}

Our model architecture is illustrated in Fig. \ref{fig2}. Our approach leverages deep AUC maximization \cite{ yuan2021robust}, a novel surrogate loss proposed for medical image classification, with self-supervised pre-training to maximize label efficiency, performance, and model trust. In our experiments, we compare the loss function against traditional cross-entropy (CE) optimization on both self-supervised and supervised models. The self-supervised model is built on the MoCo framework \cite{He_2020_CVPR} and it is pre-trained on MIMIC-CXR dataset \cite{Johnson2019}. All models are then fine-tuned on the COVIDx8B dataset \cite{Wang2020} for validation. DenseNet-121 is chosen as the backbone architecture throughout our experiments \cite{8099726}. 

\subsection{Datasets}

The MoCo model has been pre-trained on the MIMIC-CXR dataset for predicting patient deterioration \cite{sriram2021covid}. The dataset is composed of 377,110 chest radiographs \cite{Johnson2019}. As the dataset was constructed before the COVID-19 pandemic, it does not contain any positive chest X-ray samples of COVID-19.


\begin{table}[h!]
    \caption{Data split for COVIDx8B}
    \renewcommand{\arraystretch}{1.2}
    \centering
    \begin{tabular}{|c | c c c|} 
         \hline
         \textbf{Split} & \textbf{Negative} & \textbf{Positive} & \textbf{Total} \\ 
         \hline
         \textbf{Train} & 13,794 & 2,158 & 15,952 \\ 
         \textbf{Test}  & 200   & 200  & 400 \\  
         \hline
    \end{tabular}
    \label{tablecovid}
\end{table}


We perform end-to-end fine tuning on the COVIDx8B dataset \cite{Wang2020}. The latest version of COVIDx8B consists of 15,952 chest radiographs for training and 400 for testing (Table \ref{tablecovid}). Each sample in the dataset is labelled as either COVID-19 positive or negative. Stratified 5-fold cross validation is conducted on the training split during the fine-tuning stage to evaluate model performance. 

\subsection{Experiment Setups}

The DenseNet-121 model pre-trained on MIMIC-CXR using the MoCo framework has a projection dimension of 128, whereas the supervised model pre-trained on ImageNet has a projection dimension of 1,000. For end-to-end fine tuning, the parameters of the last fully connected layer of both pre-trained models are replaced and randomly initialized with a single output neuron for binary classification. We apply a sigmoid layer over the raw logits of the model to obtain a probability distribution. All input images are resized to 224x224, center cropped and normalized. Only random horizontal flipping is used for data augmentation as further augmentations were noted to provide little improvement for classification \cite{sriram2021covid}.\\    

\textbf{AUC Maximization.} We adpot a novel surrogate loss function introduced by \cite{yuan2021robust} to maximize the area under the Receiver Operating Characteristic curve. For end-to-end fine tuning, we use a learning rate of $0.1$ for all layers of the DenseNet model. Then, we optimize the network with the new surrogate loss function to maximize the AUC metric. Lastly, we train for 30 epochs while decaying the learning rate by a factor of 10 at the 15th epoch.\\   

\textbf{CE Optimization.} For standard end-to-end fine tuning, we set the learning rate at $1e-3$ for all layers of the DenseNet model. Following similar procedures to \cite{sriram2021covid}, we use cosine annealing learning rate decay to reduce the learning rate. Finally, we use the SGD optimizer on cross-entropy loss with a momentum of 0.9 and weight decay of $1e-4$ to fine tune the model for 30 epochs.\\

During each validation fold, we first compute an optimal threshold by maximizing F1-score on the validation split. Then, we save the model corresponding to the best validation accuracy. Finally, we evaluate the saved models on the unseen test split.

\section{Experimental Results}

\subsection{Supervised vs. Self-Supervised Pre-training}
We first examine the performance difference between traditional supervised pre-training on ImageNet and self-supervised contrastive pre-training on MIMIC-CXR. Tables \ref{tableprecision} and \ref{tablesensitivity} show significant improvements in the precision metric of the negative class and the sensitivity metric of the positive class for both CE optimization and AUC maximization. In medical image analysis, this improvement is key as maximizing the positive sensitivity score is necessary to lower false-negatives.\\

However, this increase in performance comes at the cost of model trust. We examine the trustworthiness of each model by calculating a trust score of the positive class. As per the methods introduced in \cite{wong2021really}, we compute a score that rewards well-placed confidence and penalizes undeserved overconfidence. In Table \ref{TableTrust}, we notice that in the case of CE optimization, supervised models are drastically more trustworthy than self-supervised models. Moreover, throughout our CE optimization experiments, we observed that self-supervised models are less confident in its correct predictions (overcautious) than its supervised counterparts.

\subsection{CE Optimization vs. AUC Maximization}

\begin{figure}[h!]
	\begin{center}
		\includegraphics[scale = 0.285]{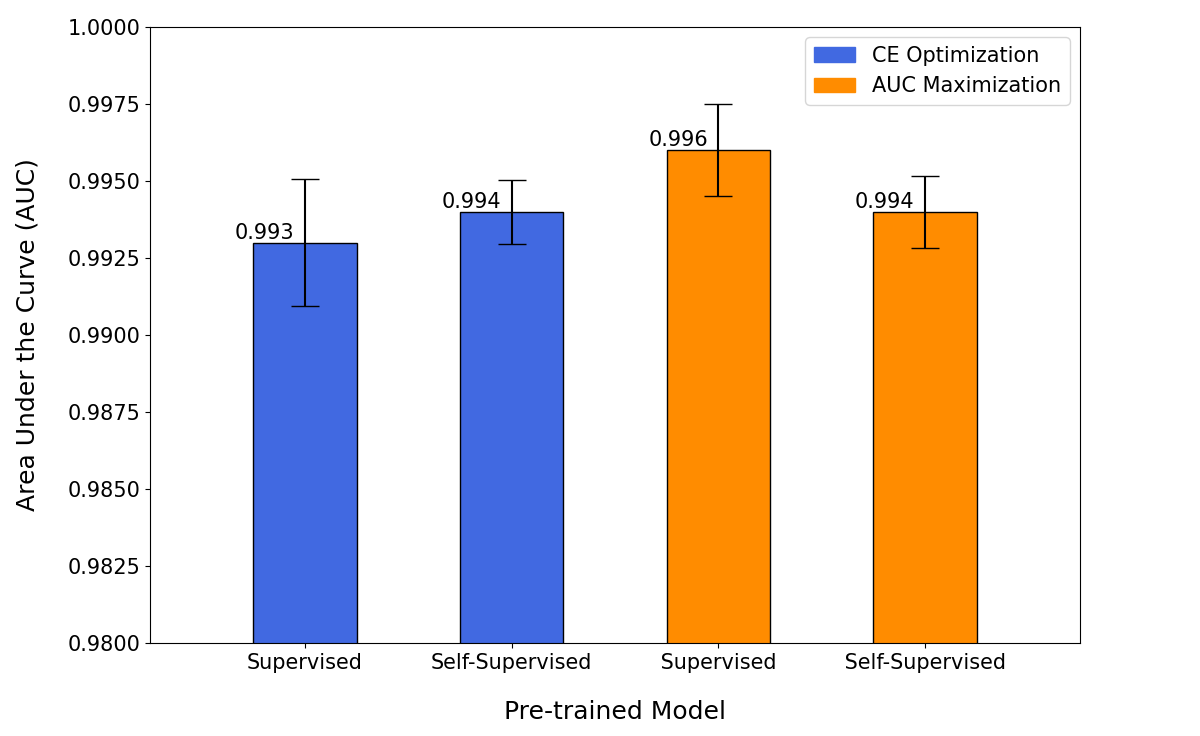}
	\end{center}
	\caption{ Area under the Receiver Operating Characteristic curve. Error bars represent the standard deviation across cross-validation runs.}
	\label{fig1}
\end{figure}

Our comparisons of CE Optimization against AUC Maximization show improvements across standard metrics as well as overall model trust-worthiness. Both Table \ref{tableprecision} and Table \ref{tablesensitivity} show improvements in the precision and sensitivity metrics regarding the supervised models. Moreover, Fig. \ref{fig1} demonstrates an increase in the AUC scores of the supervised models. When examining self-supervised models, AUC maximization still achieves performance comparable to CE optimization.\\ 

More importantly, we observe significant gains in model trust scores, especially in the context of self-supervised models. Table \ref{TableTrust} shows a near 1\% increase in supervised pre-training and a near 6\% increase in self-supervised pre-training. Moreover, when using AUC maximization, we do not see the same diminishment in model trust between supervised and self-supervised models. Therefore, unlike CE optimization, we can freely leverage AUC maximization with self-supervised pre-training to improve performance without sacrificing model trust. As shown in Tables \ref{tableprecision}, \ref{tablesensitivity} and \ref{TableTrust}, AUC maximization allows us to achieve top metrics without trading off model trust for performance. 


\begin{table}[h!]
    \caption{Precision scores on the unseen COVIDx8B test split. The best metric out of each optimization method is bolded. The best metric across methods is denoted by *.}
    \renewcommand{\arraystretch}{1.2}
    \centering
    \begin{tabular}{|c | c| c |} 
         \hline
         \textbf{Pre-trained Model} & \textbf{Negative} & \textbf{Positive} \\ 
         \hline
         Supervised (CE Opt)       & 0.8960 $\pm$ 1.6\% & 0.9956 $\pm$ 0.4\% \\
         Self-Supervised (CE Opt)  & \textbf{0.9295*} $\pm$ 1.6\% & \textbf{0.9978} $\pm$ 0.4\% \\  
         \hline
         Supervised (AUC Max)  & 0.9134 $\pm$ 1.4\% & 1.000 \\  
         Self-Supervised (AUC Max)  & \textbf{0.9251} $\pm$ 0.6\% & \textbf{1.000}* \\  
         \hline
    \end{tabular}
    \label{tableprecision}
\end{table}

\begin{table}[h!]
    \caption{Sensitivity scores on the unseen COVIDx8B test split. The best metric out of each optimization method is bolded. The best metric across methods is denoted by *.}
    \renewcommand{\arraystretch}{1.2}
    \centering
    \begin{tabular}{|c | c| c| } 
         \hline
         \textbf{Pre-trained Model} & \textbf{Negative} & \textbf{Positive} \\ 
         \hline
         Supervised (CE Opt)       & 0.9960 $\pm$ 0.3\% & 0.8840 $\pm$ 2.1\% \\
         Self-Supervised (CE Opt)  & \textbf{0.9980} $\pm$ 0.4\% & \textbf{0.9240*} $\pm$ 1.9\% \\  
         \hline
         Supervised (AUC Max)  & 1.000 & 0.9050 $\pm$ 1.7\% \\  
         Self-Supervised (AUC Max)  & \textbf{1.000}* &  \textbf{0.9190} $\pm$ 0.7\% \\
         \hline
    \end{tabular}
    \label{tablesensitivity}
\end{table}

\begin{table}[h!]
    \caption{Trust scores calculated from each experiment on the positive class. The best score overall is bolded.}
    \renewcommand{\arraystretch}{1.2}
    \centering
    \begin{tabular}{|c | c | c|}
        \hline
         \textbf{Cost Function} & \textbf{Supervised} & \textbf{Self-Supervised} \\ 
         \hline
         CE Opt      & 0.929 $\pm$ 0.9 \% & 0.879 $\pm$ 1.4 \% \\
         AUC Max      & \textbf{0.938} $\pm$ 1.0 \% & 0.937 $\pm$ 1.0 \% \\
         \hline
    \end{tabular}
    \label{TableTrust}
\end{table}

\section{Conclusion}

This work demonstrates that we no longer need to sacrifice model trust for performance. Integrating AUC maximization can produce more trustworthy and better performing models. By extending the AUC maximization paradigm \cite{yuan2021robust} to self-supervised pre-training, we showed that we can significantly improve key metrics while also maintaining model trust.\\


We expect that our study on self-supervised learning with AUC maximization will contribute to the classification of both COVID-19 and future illnesses. More often than not, we cannot afford to collect large amount of labeled samples at the onset of a pandemic. Therefore, it is important that we exploit existing data, apply effective representation learning to maximizing model performance, and gain optimal model confidence.




\section*{Acknowledgments}
We would like to thank Jianxing (Jason) Zhang for his preliminary work and code preparation that led to the completion of this work. We also acknowledge support from the Pandemic Response Challenge Program at the National Research Council of Canada.

\bibliographystyle{IEEEtran}  
\bibliography{main.bib}
\end{document}